\documentclass[conference]{IEEEtran}
\IEEEoverridecommandlockouts

\usepackage{amsmath,amssymb,amsfonts}
\usepackage{algorithmic}
\usepackage{graphicx}
\usepackage{textcomp}
\usepackage{xcolor}
\def\BibTeX{{\rm B\kern-.05em{\sc i\kern-.025em b}\kern-.08em
    T\kern-.1667em\lower.7ex\hbox{E}\kern-.125emX}}

\usepackage{bm}
\usepackage{booktabs}
\usepackage{multirow} 
\usepackage{caption}
\usepackage{hyperref}

\begin{document}

\title{Accelerating IC Thermal Simulation Data Generation via Block Krylov and Operator Action
}


\author{
    \IEEEauthorblockN{
        Hong Wang,
        Wenkai Yang,
        Jie Wang\IEEEauthorrefmark{2}\thanks{$^\dag$ Corresponding author.},
        Huanshuo Dong,
        Zijie Geng,
        Zhen Huang,\\
        Depeng Xie,
        Zhezheng Hao,
        Hande Dong
    }
    \IEEEauthorblockA{
        University of Science and Technology of China, Hefei, Anhui, China. Email: wanghong1700@mail.ustc.edu.cn
    }
}

\maketitle

\begin{abstract}
Recent advances in data-driven approaches, such as neural operators (NOs), have shown substantial efficacy in reducing the solution time for integrated circuit (IC) thermal simulations.
However, a limitation of these approaches is requiring a large amount of high-fidelity training data, such as chip parameters and temperature distributions, thereby incurring significant computational costs.
To address this challenge, we propose a novel algorithm for the generation of IC thermal simulation data, named block Krylov and operator action (BlocKOA), which simultaneously accelerates the data generation process and enhances the precision of generated data.
BlocKOA is specifically designed for IC applications. 
Initially, we use the block Krylov algorithm based on the structure of the heat equation to quickly obtain a few basic solutions.
Then we combine them to get numerous temperature distributions that satisfy the physical constraints.
Finally, we apply heat operators on these functions to determine the heat source distributions, efficiently generating precise data points.
Theoretical analysis shows that the time complexity of BlocKOA is one order lower than the existing method.
Experimental results further validate its efficiency, showing that BlocKOA achieves a 420-fold speedup in generating thermal simulation data for 5000 chips with varying physical parameters and IC structures.
Even with just 4\% of the generation time, data-driven approaches trained on the data generated by BlocKOA exhibits comparable performance to that using the existing method.
\end{abstract}


\section{Introduction and Related Work}




The temperature distribution of an integrated circuit (IC)  directly affects its performance, reliability, and lifespan~\cite{delaram2018optimal}. As a result, thermal optimization has become an essential step in the design process, involving numerous numerical thermal simulations~\cite{cao2019survey}. 
Traditionally, thermal simulations rely heavily on computationally intensive methods~\cite{sultan2019survey}. 
To reduce simulation time and enhance the efficiency of thermal optimization, recent studies have explored data-driven approaches to solve heat equations~\cite{kumar2022ml, wen2020dnn}. One prominent method involves neural operators (NOs)~\cite{li2020fourier, lu2019deeponet}, which can be trained on pre-generated datasets as surrogate models.
During practical applications, they only require a straightforward forward pass to predict  IC temperature distributions  which takes only several milliseconds~\cite{wangaro}. 
This is significantly faster than numerical partial differential equation (PDE) solvers, thermal resistance models, and other conventional algorithms.

\begin{figure*}[ht]
\begin{center}
    \centering
    \raisebox{1cm}{\includegraphics[width=0.9\linewidth]{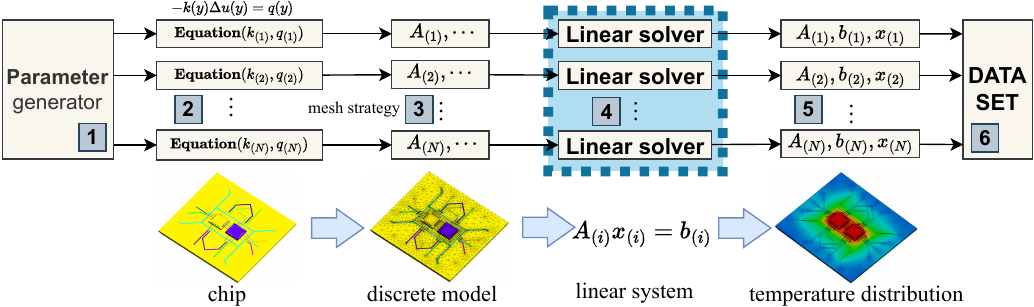}}
\vskip -0.45in
\caption{
The typical generation process of the thermal simulation dataset: 1. Produce a collection of random parameters derived from chips 2. Generate the relevant chips using these parameters 3. Discretize the chips using the FEM 4. Solve linear systems 5. Acquire solutions for the linear systems and convert them into temperature distributions 6. Compile the data into a dataset.
}
\label{figback}
\end{center}
\vskip -0.3in
\end{figure*}

However, the high computational cost of generating training datasets, which comprise chip structures and temperature distributions, presents significant challenges for these data-driven algorithms. 
First, chip applications often involve various chip types, and training a neural network for a specific type of IC requires a large volume of training data~\cite{hao2022physics}. 
For example, training DeepOHeat typically requires thousands of temperature distributions under different power and IC structures~\cite{liu2023deepoheat}. 
Dataset generation can range from several to thousands of hours.
Secondly, acquiring temperature distributions as training labels poses a further obstacle.
Data generation often relies on finite element method (FEM)~\cite{sultan2019survey}, demonstrated in Figure~\ref{figback}, involves solving large linear systems, with a high computational complexity that accounts for over 95\% of the data generation time~\cite{szabo2021finite}. 
Thirdly, solving large linear systems often involves iterative methods such as conjugate gradient (CG)~\cite{golub2013matrix}. 
However, due to the presence of termination conditions, these methods inevitably introduce errors, which can degrade the neural network's performance~\cite{lu2022comprehensive}. Increasing solution accuracy significantly raises computational costs~\cite{van2003iterative}.  
Furthermore, unlike other domains, IC heat equation parameters are closely tied to the characteristics of the chip, resulting in linear systems with specific structures~\cite{sultan2019survey}.
Existing algorithms fail to exploit these structures, resulting in redundant computations. 
Altogether, these challenges in data generation significantly hinder the practical application of data-driven algorithms in IC thermal simulations~\cite{hao2022physics}.


Several studies have achieved meaningful progress.
\cite{brandstetter2022clifford, liu2023ino} proposed architectures that preserve conservation laws to enhance data efficiency. \cite{wang2024accelerating} reduced dataset generation costs by leveraging correlations between linear systems. 
However, these advances mainly focused on optimizing the solution algorithms without fundamentally changing the generation approach. 
\cite{dong2024accelerating} introduced the operator action method to reduce the generation time, but it did not consider the specific structure of IC heat equations, limiting its direct applicability.

In this work, we introduce BlocKOA (Block Krylov and Operator Action), an efficient method for generating IC thermal simulation data. 
Initially, we use the block Krylov algorithm to simultaneously solve a small number of linear systems with different thermal parameters and quickly generate a set of temperature distributions that reflect the real IC design as basic solutions.
These basic solutions are then suitably combined to satisfy the heat equation conditions and generate IC temperature distributions. 
Finally, we perform the action of heat operators on the temperature distributions to derive other physical parameters.
The key insight behind BlocKOA is that it reduces the redundant computation in generating basic solutions via block Krylov and avoids solving linear systems through operator action.
Our contributions are as follows:
\begin{itemize}
    \item [1.] A novel dataset generation algorithm for IC thermal simulation that employs block Krylov and operator action to quickly produce large-scale data.
    \item [2.] Theoretical analysis shows BlocKOA achieves higher precision at a lower cost, ensuring accuracy and efficiency.
    \item [3.] Extensive experiments demonstrate that BlocKOA substantially reduces generation time, accelerating it by up to 420 times.
\end{itemize}

\section{Background}


\subsection{Mathematical Form of IC Thermal Simulation}


We focuses on the generation of steady-state IC thermal simulation datasets, and the equation can be expressed as~\cite{li2006ic}:
\begin{small}
\begin{equation}
- k (\bm{y}) \Delta u(\bm{y}) = q(\bm{y}).
	\label{eq:heat}
\end{equation}
\end{small}
Here, \( \Delta \) is the Laplacian; \( u \) represents the temperature distribution; \(k\) is related to IC thermal conductivity and floorplan; And \( q \) is the power density (heat generation rate)~\cite{hua2023estimation}. 






IC thermal simulations usually consider uniform boundary shapes like rectangular prisms and involve various boundary conditions~\cite{sultan2019survey}. We discuss the 3 most common types~\cite{thomas2013numerical}:
\textbf{Dirichlet}:  \( u = u_0 \), modeling contact with objects at a fixed temperature.
\textbf{Robin}: For simulating convective heat transfer at boundaries:
$k \frac{\partial u}{\partial n} + h(u - u_{\infty}) = 0$.
Here \( h \) is the convective heat transfer coefficient (HTC) and \( u_{\infty} \) is the ambient temperature.
\textbf{Mixed}: Combine Dirichlet and Robin conditions.

\subsection{Discretization for the Heat Equation}\label{Discretization}

IC thermal simulation datasets are obtained by solving the corresponding heat equation, typically using discretization methods like the finite element method (FEM)~\cite{strikwerda2004finite}.
These numerical methods transform the PDEs to linear systems~\cite{hughes2012finite}. 
Realistic thermal simulations, which involve more complex boundary conditions~\cite{delaram2018optimal}, require finer grid resolutions. 
Consequently, the matrix dimension in the resulting linear systems can increase dramatically, from \(10^3\) to \(10^7\) or more, leading to substantial computational costs in dataset generation.

Unlike other PDE datasets, the parameters in the heat equation are constrained by the specific chip structure. 
For example, the thermal conductivity parameter \( k \) determines \( \bm{A} \), but \( k \) cannot be generated randomly and usually depends on the specific chip structure. 
However, the number of publicly available chip structures is limited, leading to linear systems with the same \( \bm{A} \) in the dataset. 
Independently solving these problems with existing algorithms can cause significant redundancy. 
We introduce a block algorithm to simultaneously perform Krylov subspace iterations for linear systems with the same \( \bm{A} \). 
By avoiding redundant computations, thereby significantly reducing the computational overhead.

\subsection{Details of the Dataset}


The heat eqution is discretized on a \( N_g \times N_g \times N_g \) uniform grid \( \Omega = \{ (i_1/N_g, i_2/N_g, i_3/N_g) \mid i_1, i_2, i_3 = 0, 1, \dotsc, N_g \} \). Therefore, the dimension of the matrix \( \bm{A} \) obtained from the discretized heat equation is \( N_g^3 \).
Based on the grid \( \Omega \), we generate a dataset with features \( F_l = (k_l(\Omega), q_l(\Omega)) \) and target \( T_l = u_l(\Omega) \), where \( l = 1, 2, \dotsc, N_{\text{data}} \). 
In existing data generation methods, the solution \( u \) is obtained by solving the equation with given \( k \), \( q \), and boundary conditions. 
The function \( k \), directly related to the chip's materials and component floorplan, is limited due to the scarcity of publicly available chip designs—typically, a dataset with \( 10^3 \) samples may have only $5-100$ different chip floorplans ~\cite{wangaro}. 
In contrast, \( q \) can be generated in many ways (e.g., random constants and Gaussian random fields consistent with chip structures), leading to distinct \( u \) values for each sample~\cite{wang2024accelerating}.




\subsection{Direct Solution Method}



Existing generation typically employs direct solution method~\cite{sultan2019survey, liu2023deepoheat}. 
This method randomly generates \( q \), inputs its discretized form $\bm{b}$ into the linear system, and solves $\bm{x}$ accordingly. 
This process requires solving large linear systems independently for each heat equation, which is time-consuming and becomes a bottleneck in real application.


\section{Method}\label{Method}

\begin{figure}[t]
\begin{center}
\centerline{\includegraphics[width=1\columnwidth]{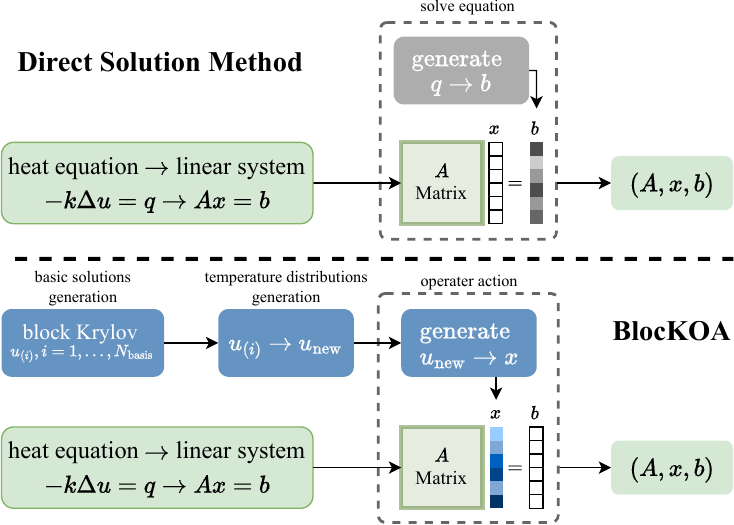}}
\caption{
Overview of the model architecture: the process of the existing direct solution method and our BlocKOA method.
}
\label{figmethod}
\end{center}
\vskip -0.45in
\end{figure}

As shown in Figure~\ref{figmethod}, unlike the direct solution method, our BlocKOA method first generates the basic solutions using the block Krylov algorithm, then combines them to produce feasible temperature distributions \( u_{\text{new}} \). 
Finally, BlocKOA inputs their discretized form $\bm{x}$ into linear systems and calculates $\bm{b}$ accordingly.
Both methods produce data that comply with their respective heat equation constraints. 
However, BlocKOA leverages the chip structure to reduce redundant computations. It also avoids the high computational costs and termination errors typically encountered in solving large linear systems.

\subsection{Basic Solutions Generation}

The BlocKOA method randomly generates a set of IC thermal parameter distributions (e.g., $N_{\text{basis}} = 50$), like the direct solution method. Subsequently, the corresponding linear systems are solved to obtain a small set of solutions $x_{(i)}$:
\begin{small}
\begin{equation}
\begin{aligned}
    - k_{(i)}(\bm{y}) \Delta u_{(i)}(\bm{y}) = q_{(i)}(\bm{y}) \rightarrow \bm{A}_{(i)}\bm{x}_{(i)}=\bm{b}_{(i)},
    \label{eq:linear equation}
\end{aligned}     
\end{equation}
\end{small}
where $i=1,\dots, N_{\text{basis}}$. If the direct solution method is used, this step requires solving \(N_{\text{basis}} \) linear systems independently.
Given the limited number of publicly available chip component floorplans (e.g., \( N_k = 5 \)), we set \( N_{\text{basis}} \) as a multiple of \( N_k \), denoted as \( \eta = {N_{\text{basis}}}/{N_k} = 10 \). This means that the resulting linear systems contain repeated matrices \( \bm{A} \):
\begin{small}
\begin{equation}
\bm{A}_{(\eta j+1)}  = \bm{A}_{(\eta j+2)} = \dots = \bm{A}_{(\eta (j+1))}, \ j = 0, \dots, N_k - 1.
\end{equation}  
\end{small}
To improve computational efficiency, we reorganize Eq. \eqref{eq:linear equation}. Then we apply the block Krylov algorithm to solve the linear systems with the same coefficient matrix $\bm{A}$ simultaneously:
\begin{small}
\begin{equation}
    \bm{A}_{(\eta j+1)} \bm{X}_{(j)} = \bm{B}_{(j)}, \quad j = 0, \dots, N_k - 1, 
\label{eq:A(i)_1}
\end{equation}
\end{small}
where $\bm{X}_{(j)} = [\bm{x}_{(\eta j+1)} \mid \bm{x}_{(\eta j+2)} \mid \dots \mid \bm{x}_{(\eta (j+1))}]$ and $\bm{B}_{(j)} = [\bm{b}_{(\eta j+1)} \mid \bm{b}_{(\eta j+2)} \mid \dots \mid \bm{b}_{(\eta (j+1))}]$.
The block algorithm (e.g., block CG) combines multiple linear systems into a block system~\cite{golub2013matrix}, effectively exploiting the shared structure of \(\bm{A}\) and the similarities in the computational process. By simultaneously handling multiple right-hand sides, it reduces the size of the required Krylov subspace compared to existing method that solve them independently, thereby decreasing computational costs.
Finally \( \bm{x}_{(i)} \) are converted into the basis solutions \( u_{(i)} \) according to the discretization form.



\subsection{Temperature Distributions Generation}

We apply a Gaussian distribution to randomly weight and normalize the previously obtained basis solutions \(u_{(i)}\) to obtain a large number of diverse solution functions that satisfy physical constraints.
Specifically, we introduce a noise element \(\epsilon\) (e.g., the normal distribution) that maintains the boundary conditions unaltered, resulting in a new temperature distribution function \(u_{\text{new}}\):
\begin{small}
\begin{align}
u_{\text{new}}(\bm{y}) = \sum_{i=1}^{N_{\text{basis}}}\alpha_{i} \cdot u_{(i)}(\bm{y})+\epsilon,& &
\alpha_{i}=\frac{\mu_i}{\sum_{j=1}^{N_{\text{basis}}}\mu_j},
\end{align}
\end{small}
where $\mu_i \sim N(0,1), \ i=1,2,...,N_{\text{basis}}$.
This method of weighting ensures that the newly formulated temperature distributions \( u_{\text{new}}\) comply with the heat equation.
The incorporation of noise helps to enhance the complexity and generalization ability of the generated dataset.  
Through this method, we are able to generate a large number of physically meaningful temperature distributions with minimal cost.

\subsection{Operator Action}

The differential operator \( -k\Delta \) in the heat equation~\eqref{eq:heat} represents a mapping within the Sobolev space. This operator maps functions to other functions, as described below:
\begin{small}
\begin{equation}
    -k(\bm{y})\Delta: \mathcal{U} \rightarrow \mathcal{Q}, \quad u (\bm{y}) \mapsto q(\bm{y}).
    \label{eq:L}    
\end{equation}
\end{small}
Here \( \mathcal{U} \) and \( \mathcal{Q} \) represent the Sobolev spaces of the temperature distribution and power density function, respectively. 
The operator action represents the action of \( -k\Delta \) on \( u \). 
We obtain \( q \) by applying the operator to the temperature distributions \(u_{\text{new}}\) generated in the previous step.

In practice, when applying the operator to a function, we discretize the differential equation. The operator \( -k\Delta \) is represented as a linear transformation \( \bm{A} \), where  \( u \) is represented by the vector \( \bm{x} \), and \( q \) is represented by the vector \( \bm{b} \).
In direct solution methods, this process is transformed into solving a large linear system by computing \( \bm{x} \) from \( \bm{A} \) and \( \bm{b} \). However, in BlocKOA, the operator action is directly represented as a matrix-vector multiplication \( \bm{A} \bm{x} \mapsto \bm{b} \), which avoids solving the linear system.
For the same problem, the computational cost of a single matrix-vector multiplication is significantly lower than that of solving the corresponding linear system, and no additional errors are introduced during the computation. Therefore,  BlocKOA offers greater speed and higher precision.

\section{Theoretical Analysis}\label{Theoretical}


\subsection{Computational Complexity Analysis}\label{Complexity}

\subsubsection{Direct Solution Method}


The primary computational cost in numerical thermal simulation arises from solving the corresponding linear systems, with the Krylov subspace algorithm (e.g., CG) being among the most commonly used approaches~\cite{sultan2019survey, ladenheim2018mta}. 
The most computationally intensive components are the matrix-vector multiplication and the orthogonalization process~\cite{liesen2013krylov}.
Assuming a matrix of dimension \( n \) and iteration count \( j = 1, 2, \dots, m \), where \( m \) denotes the final dimension of the Krylov subspace.
The complexity per iteration is primarily determined by \( O(n^2) \) for matrix-vector multiplications and \( O(j   n) \) for orthogonalization. 
The total complexity across \( m \) iterations can be approximated by \( O(m   n^2) \) for matrix-vector products and \( O(m^2   n) \) for orthogonalization, yielding an overall cost of approximately \( O(m n^2 + m^2 n) \). 
Furthermore, practical complexity is influenced by matrix sparsity.
For a dataset with \( N_{\text{data}} \) data points, the computational complexity can be  by \( O(m   n^2   N_{\text{data}} + m^2   n   N_{\text{data}}) \).

\subsubsection{BlocKOA Method}


Let \( N_{\text{basis}} \) represent the number of basis solutions, which is typically much smaller than the dataset size \( N_{\text{data}} \) (e.g., \( N_{\text{basis}} = 50 \) and \( N_{\text{data}} = 5 \times 10^3 \)). The BlocKOA method consists of 3 steps: 1. The time complexity for “Basic Solution Generation” can be expressed as \( O(m    n^2    N_{\text{basis}} + m^2    n    N_{\text{basis}}) \).
2. Constructing the temperature distributions from the basic solutions involves only matrix and vector additions, which contribute negligible computational cost.
3. The computational cost for the operator action is equivalent to a single matrix-vector multiplication, with a complexity of \( O(n^2) \) for dense matrices. Specifically, matrix sparsity affects this cost. For a dataset with \( N_{\text{data}} \) data points, this part incurs a computational cost of  \( O(n^2 N_{\text{data}}) \).

Consequently, the overall complexity of BlocKOA can be approximated by \( O(n^2    N_{\text{data}} + m    n^2    N_{\text{basis}} + m^2    n    N_{\text{basis}}) \). Compared to the direct solution method, BlocKOA is computationally advantageous because \( N_{\text{basis}} \) is significantly smaller than \( N_{\text{data}} \). 
Additionally, since \( m \) is of the same order as \( n \) (typically between \( n/20 \) to \( n/5 \) in experiments), our approach theoretically provides an approximate speedup of \( O(m) \approx O(n) \), which corresponds to an increase in speed by one order.
The block algorithm can be used to significantly reduce the number of iterations of "Basic Solution Generation" by sharing the Krylov subspace.
Typically \(m\) can be reduced to \(m/10-m/2\) in experiments. Therefore, using the block algorithm can speed up our algorithm by \(2-10\) times as a whole.


\subsection{Error Analysis}

When constructing a thermal simulation dataset, errors stem mainly from three sources: 1. PDE modeling error; 2. grid discretization error; 3. linear system error. Our analysis focuses on the third component, assuming the first two are negligible.

Iterative methods (i.e., CG) generate approximate solutions that improve with each iteration.
Let error be \( e_m = \| \bm{x}_m - \bm{x} \| \), where \( m \) is the number of iterations, \(x_m\) is the solution obtained at the \(m\)-th iteration, \(x\) is the true solution.
\( e_m \) is related to the condition number \( \kappa \) of \( \bm{A}\) and the initial error \( e_0 \)~\cite{van2003iterative}:
$e_m \leq 2 \left( \frac{\sqrt{\kappa} - 1}{\sqrt{\kappa
} + 1} \right)^m e_0$. 
Reducing \( e_m \) requires increasing \( m \), that will increase the computational cost.

It is generally required that the error of the dataset be significantly lower than that of the data-driven algorithm.
For instance, in algorithms like \cite{liu2023deepoheat}, a final error range of $1\mathrm{E}{-2}$ to $1\mathrm{E}{-5}$ is typical, ideally with relative dataset errors is lower than $1\mathrm{E}{-7}$ to maintain training accuracy. For direct solution method, higher accuracy requires more iterations.

In BlocKOA, applying the operator to the generated temperature distributions essentially involves matrix-vector multiplication. 
The precision of this operation is governed by the machine epsilon of floating-point operations, generally yielding error is lower than $1\mathrm{E}{-16}$.
Achieving this high level of precision is impossible with the direct solution method.






\section{Experiment}

\begin{table*}[ht]
\vskip -0.2in
\centering
\caption{
Comparison of data generation time (in seconds) for BlocKOA and CG under different IC scenarios. BlocKOA achieves a machine precision error of $1\mathrm{E}{-16}$. The first row lists IC boundary conditions, while TIME1 and TIME2 in the third row represent the total data generation time and operator action time, respectively. Other parameters in the third row represent CG errors (relative residual norm). Dim represents the matrix dimension.
}
\label{tab:time_and_accuracy}
\vskip -0.08in
\fontsize{6}{8}\selectfont  
\begin{sc}
\renewcommand{\arraystretch}{1}
\setlength{\tabcolsep}{4pt}
\begin{tabular}{@{}cccccccccccccccccccccccc@{}}
\toprule
\multirow{3}{*}{Dim} & \multicolumn{7}{c}{\textbf{Dirichlet}} & \quad & \multicolumn{7}{c}{\textbf{Robin}} & \quad & \multicolumn{7}{c}{\textbf{Mixed}} \\ \cline{2-8} \cline{10-16} \cline{18-24}
& \multicolumn{2}{c}{\textbf{BlocKOA}} & \quad & \multicolumn{4}{c}{\textbf{CG}} & \quad & \multicolumn{2}{c}{\textbf{BlocKOA}} & \quad & \multicolumn{4}{c}{\textbf{CG}} & \quad & \multicolumn{2}{c}{\textbf{BlocKOA}} & \quad & \multicolumn{4}{c}{\textbf{CG}} \\ \cline{2-3} \cline{5-8} \cline{10-11} \cline{13-16} \cline{18-19} \cline{21-24}
& time1 & time2 & \quad & 1e-3 & 1e-5 & 1e-7 & 1e-9 & \quad & time1 & time2 & \quad & 1e-3 & 1e-5 & 1e-7 & 1e-9 & \quad & time1 & time2 & \quad & 1e-3 & 1e-5 & 1e-7 & 1e-9 \\ \hline
$6.0\times10^4$  & 2.1e1 & 1.4e0 & \quad & 2.2e3 & 3.5e3 & 4.5e3 & 5.7e3 & \quad & 2.4e1 & 1.6e0 & \quad & 2.2e3 & 3.8e3 & 5.4e3 & 5.9e3 & \quad & 2.0e1 & 1.3e0 & \quad & 2.4e3 & 3.8e3 & 5.7e3 & 7.4e3 \\
$8.5\times10^4$  & 3.8e1 & 2.1e0 & \quad & 3.3e3 & 5.6e3 & 7.8e3 & 1.0e4 & \quad & 4.5e1 & 3.0e0 & \quad & 3.9e3 & 6.2e3 & 7.5e3 & 9.0e3 & \quad & 3.5e1 & 1.8e0 & \quad & 3.8e3 & 6.3e3 & 7.9e3 & 9.5e3 \\
$1.0\times10^5$ & 5.2e1 & 2.7e1 & \quad & 5.1e3 & 8.1e3 & 1.2e4 & 1.6e4 & \quad & 5.6e1 & 3.2e0 & \quad & 5.8e3 & 9.6e3 & 1.2e4 & 1.5e4 & \quad & 5.0e1 & 3.2e0 & \quad & 6.1e3 & 9.7e3 & 1.3e4 & 1.5e4 \\
$1.3\times10^5$ & 7.0e1 & 2.3e1 & \quad & 7.5e3 & 1.3e4 & 1.8e4 & 2.2e4 & \quad & 8.5e1 & 4.1e0 & \quad & 8.7e3 & 1.5e4 & 1.9e4 & 2.4e4 & \quad & 6.8e1 & 3.8e0 & \quad & 9.9e3 & 1.5e4 & 2.3e4 & 2.9e4 \\
$2.7\times10^5$ & 2.7e2 & 4.2e1 & \quad & 3.4e4 & 5.1e4 & 6.6e4 & 8.4e4 & \quad & 2.9e2 & 6.1e0 & \quad & 3.8e4 & 6.0e4 & 8.5e4 & 1.0e5 & \quad & 2.7e2 & 7.0e0 & \quad & 4.3e4 & 6.5e4 & 9.1e4 & 1.1e5 \\
\bottomrule
\end{tabular}
\end{sc}
\end{table*}

\begin{table*}[h]
\centering
\caption{
Comparison of data generation time (in seconds) and training results (RMSE) across different models (test data is generated by CG). The first row lists the IC boundary conditions, the second row lists the training data generation times and neural operators, the first column list the methods used for training data generation and $N_{\text{data}}$ is in brackets.
}
\label{tab:accuracy}
\fontsize{7}{9}\selectfont  
\begin{center}
\vskip -0.15in
\begin{sc}
\begin{tabular}{@{}lccccccccccc@{}}
\toprule
\multicolumn{1}{l}{\multirow{2}{*}{{Method}}}& \multicolumn{3}{c}{\textbf{Dirichlet}} & \quad & \multicolumn{3}{c}{\textbf{Robin}} & \quad & \multicolumn{3}{c}{\textbf{Mixed}} \\ \cline{2-4} \cline{6-8} \cline{10-12}
\multicolumn{1}{c}{} & TIME$(s)$ & FNO & DeepONet & \quad  & TIME$(s)$ & FNO & DeepONet & \quad & TIME$(s)$ & FNO & DeepONet \\[0pt] \midrule
CG (500) & 1.64e3 & 2.41e-3 & 6.44e-4 & \quad & 1.50e3 & 8.07e-3 & 1.20e-2 & \quad & 1.54e3 & 8.52e-3 & 8.01e-3 \\
CG (1000) & 3.28e3 & 2.13e-3 & 3.41e-4 & \quad & 3.01e3 & 7.39e-3 & 8.31e-2 & \quad & 3.08e3 & 7.32e-3 & 6.84e-3 \\
BlocKOA (500) & 5.24e1 & 5.61e-3 & 1.68e-3 & \quad & 5.62e1 & 1.23e-2 & 1.35e-2 & \quad & 4.85e1 & 1.49e-2 & 1.48e-2 \\
BlocKOA (2000) & 5.33e1 & 2.17e-3 & 3.64e-4 & \quad & 5.70e1 & 6.38e-3 & 7.59e-3 & \quad & 4.93e1 & 8.31e-3 & 5.49e-3 \\
BlocKOA (5000)& 5.44e1 & 1.85e-3 & 3.12e-4 & \quad & 5.82e1 & 6.14e-3 & 6.74e-3 & \quad & 5.04e1 & 7.12e-3 & 5.38e-3 \\
\bottomrule
\end{tabular}
\end{sc}
\end{center}
\vskip -0.2in
\end{table*}

\subsection{Experimental Details}






\subsubsection{IC Problem Details}


We consider the thermal simulation for 3D-ICs. Specifically, we discuss a rectangular chip measuring $10\,\text{mm} \times 10\,\text{mm} \times 0.51\,\text{mm}$. 
The chip consists of three device layers, each of $0.15\,\text{mm}$ thickness, namely: 1. the topmost core layer; 2. two L2 Cache layers with identical structures. Beneath each device layer lies a $ 0.02\,\text{mm}$-thick thermal interface material (TIM) layer (comprising bumps, redistribution layers, pads, and underfills). 
The address and data buses between the L2 Cache and core layer modules are interconnected using through-silicon vias, which are made of copper.
The thermal conductivity values used in the simulation are $150\,\text{W}/\text{mK}$ for silicon, $413\,\text{W}/\text{mK}$ for copper, and an equivalent thermal conductivity of $40\,\text{W}/\text{mK}$ for the TIM layer.
In the core layer, we place $156$ active blocks. 
$6$ of them are randomly selected as high-power modules, with power densities randomly assigned between $3$ and $6\,\text{W}/\text{mm}^2$. 
The remaining $150$ modules have power densities randomly assigned between $0.5$ and $1\,\text{W}/\text{mm}^2$. 
Each L2 Cache layer consists of two L2 Caches, with power densities randomly assigned between $0.02$ and $0.04\,\text{W}/\text{mm}^2$.

We consider 3 boundary conditions:
1. {Dirichlet}: All boundary temperatures are fixed at $50^\circ\text{C}$.
2. {Robin}: All boundaries have a HTC of $3 \times10^4$ $\text{W}/\text{m}^2\text{K}$.
3. {Mixed}: The top and bottom surfaces are Robin boundaries with HTC at $3 \times10^4$ $\text{W}/\text{m}^2\text{K}$, and the other four sides are fixed at $50^\circ\text{C}$.
We consider 5 mesh resolutions, corresponding to matrix dimensions ranging from $6.0\times10^4$ to $2.7\times10^5$. These meshes are obtained by dividing them using the open-source tool Gmesh~\cite{geuzaine2009gmsh}.
%
Each dataset considers $N_k$ (e.g. $N_k = 5$) different component floorplans, which are manually designed to mimic real chips rather than being randomly generated. 
For each component floorplan, several reasonable power distributions are generated using the above random strategy. 
Then we use the open-source finite element IC thermal simulation tool, manchester thermal analyzer (MTA)~\cite{ladenheim2018mta}, to obtain the linear systems. Finally, we use the professional linear system solver library PETSc to obtain the chip temperature distributions~\cite{petsc-web-page}.

\subsubsection{Baselines Details}


The main time expense of the existing direct solution method is solving linear systems composed of large sparse symmetric matrices~\cite{hughes2012finite}. 
In all experiments, We do not consider the time of the finite element numerical discretization process. 
We use the direct solution method based on the CG algorithm as our baseline, utilizing PETSc 3.19~\cite{petsc-web-page}. 
PETSc is a state-of-the-art linear solver library and serves as the underlying solver for many professional thermal simulation software packages (e.g., DEAL.II~\cite{arndt2020deal}, FEniCS~\cite{alnaes2015fenics}, OpenFOAM~\cite{jasak2009openfoam}, MTA~\cite{hughes2012finite}).


\subsubsection{Experimental Environment}

The data generation process is performed on an Intel i7-13700KF CPU, and the neural operator training is performed on a RTX 3090 GPU.

\subsection{Comparative Experiments with Direct Solution Method}\label{exp1}

We tested the data generation time for generating $N_{\text{data}} = 5\times10^3$  data points at different accuracies as shown in Table~\ref{tab:time_and_accuracy}. 
BlocKOA method consistently demonstrates remarkable acceleration compared to the CG method across datasets of all experiments.
We set the number of generated basis solutions \( N_{\text{basis}} = 50 \) and the number of different chip component floorplans \(N_{k} = 5\). The noise \( \epsilon \) is uniform randomly assigned within the range of $-0.01$ to $0.01$ at each grid point.

First, the experimental results show that compared to the CG method, the BlocKOA method achieves a speedup of approximately $420$ times in terms of the total time, while the time for operator actions can be accelerated by up to approximately $1.7\times10^4$ times for the matrix dimension of $2.7\times10^5$.
The data generation time in BlocKOA is divided into two components: basis solutions generation and operator action. 
Experimental results show that compared to the total time (TIME1), the operator action time (TIME2) is negligible, with basis solutions generation being the main part of BlocKOA's generation time. 
This is because operator action essentially involves a single matrix-vector multiplication, which has a minimal computational cost.

Second, during the CG algorithm's process of solving linear systems, as the accuracy requirement increases, the solution time significantly increases.
For instance, when the accuracy of the CG algorithm is improved from $1\mathrm{E}{-3}$ to $1\mathrm{E}{-9}$, the time increases by 160\% to 220\%.
In contrast, our BlocKOA method achieves a precision of $1\mathrm{E}{-16}$. 
This indicates that improving the accuracy of the CG algorithm comes with expensive computational costs, while our algorithm guarantees data accuracy at machine precision through operator actions. 


Moreover, as the matrix dimensions increase, the acceleration ratio of BlocKOA compared to CG grows steadily. 
For instance, under the Dirichlet boundary condition, the total time speedup rises from approximately $370$ to $420$ when the dimension increases from  $6.0\times10^4$ to $2.7\times10^5$. 
This is due to BlocKOA's computational complexity, which is one order lower, \(O(n)\), compared to the direct solution method, where \(n\) denotes the matrix dimension. 
This further supports the theoretical analysis in Section~\ref{Complexity}, highlighting the effectiveness of our method in high resolution.

Notably, the time required by BlocKOA to generate different quantities \( N_{\text{data}}\)  of training instances remains relatively constant. 
The primary computational cost of BlocKOA lies in generating the basis solutions. 
This time depends on the number  \( N_{\text{basis}}\) of basis solutions generated and is independent of \( N_{\text{data}}\). 
This characteristic implies that, given a fixed \( N_{\text{basis}}\), BlocKOA can generate an arbitrarily large amount of training data at minimal additional cost.

\begin{table*}[t]
\vskip -0.2in
\centering
\caption{
Comparison of data generation time (in seconds) between BlocKOA and CG across different number of generated data \( N_{\text{data}}\). The first row lists \( N_{\text{data}}\), and the first column lists the methods used for data generation.
}
\label{tab:N_data}
\begin{center}
\vskip -0.15in
\fontsize{7}{9}\selectfont  
\begin{sc}
\renewcommand{\arraystretch}{1}
\setlength{\tabcolsep}{7pt}
\begin{tabular}{@{}lccccccccccc@{}}
\toprule
$N_{data}$& 100 & 200 & 500 & 1000 & 2000 & 3000 & 6000 & 8000 & 10000 & 20000 \\
\midrule
CG & 3.08e2 & 6.17e2 & 1.54e3 & 3.08e3 & 6.17e3 & 9.25e3 & 1.85e4 & 2.47e4 & 3.08e4 & 6.17e4 \\
BlocKOA & 4.82e1 & 4.82e1 & 4.85e1 & 4.93e1 & 4.93e1 & 4.99e1 & 5.04e1 & 5.10e1 & 5.23e1 & 5.90e1 \\
\bottomrule
\end{tabular}
\end{sc}
\end{center}
\vskip -0.15in
\end{table*}

\begin{table}[h]
\centering
\caption{
Performance comparison of BlocKOA under varying number of generated basis solutions $N_\text{basis}$. The first row lists $N_\text{basis}$, the first column lists evaluation metrics: data generation time (seconds) and neural operator error (RMSE), and the second column presents baseline results from CG.
}
\label{tab:N_basis}
\begin{center}
\fontsize{7}{9}\selectfont   
\vskip -0.15in
\begin{sc}
\renewcommand{\arraystretch}{1}
\setlength{\tabcolsep}{7pt}
\resizebox{0.5\textwidth}{!}{%
\begin{tabular}{@{}lccccccc@{}}
\toprule
& CG & $10$ & $20$ & $30$ & $50$ & $80$ & $100$ \\
\midrule
TIME (s) & 3.08e3 & 9.98e0 & 2.02e1 & 3.03e1 & 5.04e1 & 8.08e1 & 1.02e2 \\
FNO & 7.32e-3 & 5.38e-1 & 4.34e-2 & 9.92e-3 & 7.12e-3 & 7.10e-3 & 7.07e-3 \\
DeepONet & 6.84e-3 & 2.36e-1 & 3.29e-2 & 6.42e-3 & 5.38e-3 & 5.32e-3 & 5.28e-3 \\
\bottomrule
\end{tabular}
}
\end{sc}
\end{center}
\vskip -0.1in
\end{table}

\begin{table}[h]
\centering
\caption{
Performance comparison of BlocKOA method under varying $N_k$. The first row lists $N_k$, the first column lists evaluation metrics: data generation time (seconds) and neural operator performance (RMSE), and the second column presents baseline results from CG.
}
\label{tab:N_k}
\begin{center}
\fontsize{7}{9}\selectfont   
\vskip -0.15in
\begin{sc}
\renewcommand{\arraystretch}{1}
\setlength{\tabcolsep}{7pt}
\begin{tabular}{@{}lccccc@{}}
\toprule
& CG & $N_k=1$ & $N_k=2$ & $N_k=5$ & $N_k=10$ \\
\midrule
TIME (s) & 3.08e3 & 3.88e1 & 4.29e1 & 5.04e1 & 8.39e1 \\
FNO & 7.32e-3 & 4.73e-1 & 3.24e-2 & 7.12e-3 & 7.08e-3 \\
DeepONet & 6.84e-3 & 2.51e-1 & 2.75e-2 & 5.38e-3 & 5.23e-3 \\
\bottomrule
\end{tabular}
\end{sc}
\end{center}
\vskip -0.2in
\end{table}

\subsection{Data Validity Experiments}






In these experiments, to validate the effectiveness of the data generated by BlocKOA, we focused on testing two widely recognized and extensively used NOs for IC thermal simulation: 1. FNO \cite{li2020fourier, wangaro},
2. DeepONet \cite{lu2019deeponet, liu2023deepoheat}.  
We set the number of generated basis solutions \( N_{\text{basis}} = 50 \) and the number of different chip component floorplans \(N_{k} = 5\). The noise \( \epsilon \) is uniform randomly assigned within the range of $-0.01$ to $0.01$ at each grid point.
Both models were evaluated using matrices of dimension \(1.0 \times 10^5\), and neural operator test data is generated by CG. The CG sets the accuracy to $1\mathrm{E}{-9}$ (relative residual norm).
The results are shown in Table~\ref{tab:accuracy}.




The generation time of BlocKOA is significantly lower than that of the direct solution method with the CG method for all experiments. For instance, BlocKOA's computation time for generating 5000 data points is approximately $\frac{1}{50}$ of CG's time for generating 1000 data points.
For fair comparison, all NOs test sets in our experiments were generated using the CG method. Consequently, when using datasets of equal size, models trained on BlocKOA-generated data initially showed slightly inferior performance compared to those trained on CG-generated data. However, BlocKOA's superior generation speed enables the creation of substantially larger datasets, which ultimately yield better model performance.
A compelling example under mixed boundary conditions shows that BlocKOA can generate $10\times$ more data points in just $\frac{1}{25}$ of CG's computation time, while achieving significantly lower training errors (17\% reduction for FNO and 32\% reduction for DeepONet). These results robustly validate the effectiveness of our data generation method.

\subsection{Parameter Analysis Experiments}
This section examines the impact of four critical algorithmic parameters on experimental results: 1. $N_\text{data}$: The number of generated data;
2. $N_k$: The number of distinct chip component floorplans;
3. $N_\text{basis}$: The number of generated basis solutions;
4. $\epsilon$: Noise element generation method.
All experiments address chip thermal simulation problems with mixed boundary conditions, employing a matrix dimension of $1.0\times10^5$. For consistent evaluation, we generate all neural operator test datasets using the CG method with $N_k = 5$. CG sets the accuracy to $1\mathrm{E}{-9}$ (relative residual norm).


\subsubsection{Analysis of $N_\text{data}$}

We set $N_\text{basis} = 50$, $N_k = 5$, with \(\epsilon\) uniformly randomized within $(-0.01, 0.01)$. The experimental results are presented in Table \ref{tab:N_data}.
First, regardless of the $N_\text{data}$ value, BlocKOA demonstrates significantly lower computational time compared to CG. 
Second, as $N_\text{data}$ increases, BlocKOA's advantage becomes more pronounced. For instance, when $N_\text{data}=100$, CG requires 6.4 times more computation time than BlocKOA. This ratio escalates to 1000 times when $N_\text{data}=20000$. This behavior occurs because the CG method solves each problem independently, resulting in linear growth of computation time with increasing $N_\text{data}$. In contrast, BlocKOA's primary computational overhead lies in the basis solution generation phase, which is independent of $N_\text{data}$. $N_\text{data}$ only affects the operator action phase, causing merely marginal increases in computation time.

\subsubsection{Analysis of $N_{k}$}

We set BlocKOA with $N_\text{data} = 5000$ and CG with $N_\text{data} = 2000$, while maintaining $N_\text{basis} = 50$ and \(\epsilon\) uniformly randomized within $(-0.01, 0.01)$. Results are shown in Table \ref{tab:N_k}.
First, $N_k$ directly affects BlocKOA's solution time, with larger values leading to longer computation. This occurs because with fixed $N_\text{basis}$, increasing $N_k$ reduces the number of linear systems solved simultaneously in the block method, thereby decreasing the exploitable redundancy and increasing computation time.
Second, $N_k$ influences the quality of BlocKOA-generated data. Test dataset was generated using CG with $N_k=5$. Results show that overly small $N_k$ values yield lower training accuracy, while values exceeding 5 produce nearly identical training outcomes. This aligns with expectations: insufficient $N_k$ limits floorplan diversity, while excessive values introduce irrelevant information.

\begin{table}
\centering
\caption{
Performance comparison of BlocKOA under different noise $\epsilon$ generation method. The first row lists $\epsilon$ generation methods, the first column lists evaluation metrics: data generation time (seconds) and neural operator error (RMSE), the second column presents results from CG.
}
\label{tab:noise}
\begin{center}
\fontsize{6}{9}\selectfont  
\vskip -0.15in
\begin{sc}
\resizebox{0.5\textwidth}{!}{%
\begin{tabular}{@{}lc|c|ccc|cc@{}}
\toprule
 & \multirow{2}{*}{{CG}} & \multirow{2}{*}{{No Noise}} & \multicolumn{3}{c|}{Uniform Random} & \multicolumn{2}{c}{Gaussian Random} \\
 & & & (-0.005,0.005) & (-0.01,0.01) & (-0.02,0.02) & $\sigma=0.002$ & $\sigma=0.01$ \\
\midrule
TIME (s) & 3.08e3 & 5.03e1 & 5.03e1 & 5.04e1 & 5.04e1 & 5.03e1 & 5.05e1 \\
FNO & 7.32e-3 & 7.97e-3 & 7.38e-3 & 7.12e-3 & 7.32e-3 & 7.39e-3 & 7.08e-3 \\
DeepONet & 6.84e-3 & 7.29e-3 & 6.18e-3 & 5.38e-3 & 6.46e-3 & 6.31e-3 & 5.32e-3 \\
\bottomrule
\end{tabular}
}
\end{sc}
\end{center}
\vskip -0.1in
\end{table}

\begin{table}
\centering
\caption{
Comparison of dataset generation time (in seconds) with different BlocKOA settings.}
\label{tab:Ablation Result}
\begin{center}
\fontsize{6}{8}\selectfont  
\vskip -0.15in
\begin{sc}
\begin{tabular}{@{}lcccc@{}}
\toprule
& BlocKOA & w/o block krylov & w/o operator action & w/o all \\ 
\midrule
time ($s$) & 5.20e1 & 1.25e2 & 6.83e3 & 1.64e4 \\
\bottomrule
\end{tabular}
\end{sc}
\end{center}
\vskip -0.2in
\end{table}


\subsubsection{Analysis of $N_\text{basis}$}
We set BlocKOA with $N_\text{data} = 5000$ and CG with $N_\text{data} = 2000$, while maintaining $N_k = 5$ and \(\epsilon\) uniformly randomized within $(-0.01, 0.01)$. Results are presented in Table \ref{tab:N_basis}.
First, BlocKOA's data generation time increases linearly with $N_\text{basis}$, consistent with our theoretical analysis in Section~\ref{Complexity}. The primary computational cost resides in the basis solution generation phase, which scales directly with $N_\text{basis}$.
Second, $N_\text{basis}$ affects data quality. Insufficient basis solutions reduce data diversity and problem representation capability, ultimately degrading model performance. The impact becomes negligible when $N_\text{basis}>50$, indicating adequate diversity has been achieved.

\subsubsection{Analysis of $\epsilon$}
We set BlocKOA with $N_\text{data} = 5000$, CG with $N_\text{data} = 2000$, $N_\text{basis} = 50$, and $N_k = 5$. Results are shown in Table \ref{tab:noise}, where ``No noise" denotes absence of additive noise, and Gaussian random represents zero-mean normal distribution with standard deviation $\sigma$.
Results demonstrate that appropriate noise addition improves the training effectiveness of BlocKOA-generated data. However, excessive noise proves detrimental.

\subsection{Ablation Experiments}


As shown in Table~\ref{tab:Ablation Result}, we consider the dataset generation time of the BlocKOA method after removing different modules. We set the number of generated basis solutions \( N_{\text{basis}} = 50 \) and the number of different chip designs \(N_{k} = 5\). The noise \( \epsilon \) is randomly assigned within the range of $-0.01$ to $0.01$ at each grid point. The matrix dimension is \(1.0 \times 10^5\), \(N_{\text{data}} = 5 \times 10^3\), and the boundary condition is Dirichlet.  
We replace block CG with standard CG for ``w/o block Krylov". 
The block algorithm achieves a $2.4$ times speedup in computation (``w/o block Krylov" is approximately equivalent to \cite{dong2024accelerating}). 
Additionally, our experiments reveal that the average number of iterations for CG is $980$, whereas block CG requires only $224$ iterations. This suggests that this acceleration stems from the block algorithm’s shared Krylov subspace, which minimizes redundant computations.
Moreover, BlocKOA is $130$ times faster than the ``w/o operator action".
In summary, these experiments demonstrate the critical importance of block Krylov and operator action in the efficiency of BlocKOA.

\section{Conclusion}



This paper presents the BlocKOA algorithm. To our knowledge, this is the first attempt to accelerate IC thermal simulation dataset generation. 
The BlocKOA ensures speed and accuracy, alleviating a significant obstacle to the development of data-driven algorithms in the field of IC thermal simulation.





\newpage

\newpage

\bibliographystyle{IEEEtran}
\bibliography{IEEEabrv,IEEEbib}

\end{document}